\documentclass[twoside,11pt]{article}

\usepackage{jmlr2e}

\usepackage{graphicx}
\usepackage{float}
\usepackage{multirow}
\usepackage{adjustbox}
\usepackage[title]{appendix}

\usepackage[symbol]{footmisc}

\jmlrheading{1}{2021}{1-48}{4/00}{10/00}{Ryan Nguyen, Shubhendu Kumar Singh, and Rahul Rai}

\ShortHeadings{Fuzzy Generative Adversarial Networks}{Nguyen, Singh, and Rai}

\begin{document}
\title{Fuzzy Generative Adversarial Networks}

\author{\name Ryan Nguyen \email rnguye2@clemson.edu \\
       \addr Department of Automotive Engineering\\
       Clemson University\\
       Greenville, SC 29616, USA
       \AND
       \name Shubhendu Kumar Singh \email shubhes@clemson.edu \\
       \addr Department of Automotive Engineering\\
       Clemson University\\
       Greenville, SC 29616, USA
       \AND
       \name Rahul Rai\footnote[1]{Corresponding author}
       \email rrai@clemson.edu \\
       \addr Department of Automotive Engineering\\
       Clemson University\\
       Greenville, SC 29616, USA}
\editor{}
\footnotetext{*Corresponding author}
\maketitle
\begin{abstract}
Generative Adversarial Networks (GANs) are well-known tools for data generation and semi-supervised classification. GANs, with less labeled data, outperform Deep Neural Networks (DNNs) and Convolutional Neural Networks (CNNs) in classification across various tasks, this shows promise for developing GANs capable of trespassing into the domain of semi-supervised regression. However, developing GANs for regression introduce two major challenges: (1) inherent instability in the GAN formulation and (2) performing regression and achieving stability simultaneously. This paper introduces techniques that show improvement in the GANs' regression capability through mean absolute error (MAE) and mean squared error (MSE). We bake a differentiable fuzzy logic system at multiple locations in a GAN because fuzzy logic systems have demonstrated high efficacy in classification and regression settings.  The fuzzy logic takes the output of either or both the generator and the discriminator to either or both predict the output, $y$, and evaluate the generator's performance. We outline the results of applying the fuzzy logic system to CGAN and summarize each approach's efficacy. This paper shows that adding a fuzzy logic layer can enhance GAN's ability to perform regression; the most desirable injection location is problem-specific, and we show this through experiments over various datasets. Besides, we demonstrate empirically that the fuzzy-infused GAN is competitive with DNNs.

\end{abstract}

\section{Introduction}
A majority of the engineering and applied science problems fall into the class of regression instead of classification. Regression continues to be the Achilles heel of the data-driven predictions, especially when it comes down to application domains  where real-world data is scarce. 
In regression, the true output, $y \in R$, is a continuous and stochastic function of the actual input, $x \in R$. Deep Neural Networks (DNNs), Recurrent Neural Networks (RNNs), and Long Short Term Memory Networks (LSTMs) have demonstrated some of the best results in regression; however, the results are still unsatisfactory. They require a lot of labeled data to train and are prone to overfitting. These reasons motivate a pivot towards new techniques, and we propose the pivot should point at generative adversarial networks (GANs). Generative models, mathematical frameworks designed to generate real data from a distribution, have garnered significant research attention for semi-supervised learning tasks. GANs are powerful generative models that have shown exceptional performance on semi-supervised classification. The research in improving the efficacy of GANs for semi-supervised regression is rapidly increasing. If GANs can achieve high accuracy (as shown in classification problems) for regression problems while reducing the amount of labeled data, the utility of GANs could be even more widespread. Currently, it is important to note, the performance of GANs on regression has significant scope for improvement.

Besides NNs, other techniques have garnered success in performing regression such as fuzzy logic models. Fuzzy logic is a generalization of boolean logic, where the value of truth is measured as $w \in [0,1]$ rather than a binary output of 0 or 1. T-norms and t-conorms, respectively, represent the statements for conjunction and disjunction. Subsequently, fuzzy implications are generated by following a set of axioms by which implications abide; these axioms are outlined in Section \ref{FuzzyImp}. Fuzzy logic has been shown to improve the evaluation of generative models. In these evaluation methods, fuzzy logic is used to determine the relative truth value of a generated image based on the probability of certain features existing. The authors choose features present in the real data and should be present in the generated data; the algorithm assigns a probability of those features existing. The probability of the existing features is used to determine the value of the truth of the generated data. This approach was proven to evaluate the generated data more accurately. The evaluation is on the generated images and text (\cite{niedermeier2020improving},  \cite{ltifi2018fuzzy}).  

\subsection{Challenges of GAN Models for Regression}
\label{Challenges}
GANs pose challenges that differ from traditional deep learning techniques. These issues can be ascribed to the game played between the discriminator and generator to achieve optimality. We will refer to this game as the GAN game throughout this paper, and in the current sub-section, we define the game and discuss the potential difficulties faced by the players, i.e., the discriminator and the generator.

The GAN game is a repeated two-player zero-sum game, where either the generator or the discriminator plays first. The original formulation of the minimax equation that governs the GAN game is shown below:

\begin{equation}
    \min_D \max_G V(D, G) = E_{x \sim p_{data}} [\log (D(x))] + E_{z \sim p_{z}} [\log (1 - D(G(z)))].
    \label{eq: minimax}
\end{equation}

$x$ is the input to the discriminator, $p_{data}$ is the distribution from which $x$ is drawn, $D(x)$ is the probability that the input is either generated or the real one.  $z$ is the input to the generators, $p_z$ is the distribution with which $z$ is drawn from, and $G(z)$ is the output of the generator (\cite{goodfellow2014generative}).  The divergence metric corresponding to this game is the Jensen-Shannon divergence (JS-divergence).  The GAN game is inherently unstable, one theory for the instabilities (i.e., mode collapse and non-convergence) concludes that the actual data and model-generated data distributions are disjoint or lying on low-dimensional manifolds.  Therefore, strong distance measures like KL-divergence and JS-divergence will max out instead of providing the generator with proper gradients to learn (\cite{arjovsky2017wasserstein}). \cite{arjovsky2017wasserstein} address these issues in their architecture: Wasserstein GAN (WGAN). The authors use a weaker distance measure, the Wasserstein metric, also known as the earth mover's distance, showing that the instabilities and mode collapse problems are ameliorate. Although effective, it does not universally solve the convergence issues with GANs (\cite{arjovsky2017wasserstein}, \cite{salimans2016improved}).  However, there are many other variants of GAN that use different divergence metrics and \cite{ge2018fictitious} discuss some of the different variants. Fictious GANs (f-GANs) are a meta-algorithm in which the discriminator is trained on the mixed outputs from a series of trained generators. According to the authors, it is an easy algorithm to implement. They develop proof by showing that fictitious play converges.  However, this meta-algorithm only works when there is a single Nash equilibrium, which we'll explain below, is not necessarily true by Sion's theorem. The algorithm works with any existing variant and has been shown to improve the results of WGAN (\cite{ge2018fictitious}).

However, another way to develop reasoning for stability issues can be derived from the contradiction of Sion's theorem, stated below:

\begin{theorem}
If $\Phi \subset \mathbb{R}^m$, $\Theta \subset \mathbb{R}^n$ such that they are compact and convex sets, and the function $V:\Phi \times \Theta \rightarrow \mathbb{R}$ is upper continuous and quasi-convex in its first argument and lower continuous and quasi-concave in its second, then we have:
\begin{equation}
    \min_{\phi \in \Phi} \max_{\theta \in \Theta} V(\phi, \theta) =  \max_{\theta \in \Theta} \min_{\phi \in \Phi} V(\phi, \theta)
\end{equation}.
\end{theorem}

In other words, Sion's theorem states that equilibrium is guaranteed to exist if the conditions mentioned above are met. In the GAN game, $\Phi$ is the space of parameters, $\theta_D$, that can be composed to define the discriminator and, $\Theta$ is the space of parameters, $\theta_G$, that can be composed to define the generator. So this is optimization in the parameter space. Consequently, the function, $V$, is not necessarily quasi-convex in its first argument and quasi-concave in its second argument, so GANs may have multiple undesirable local equilibria for specific conditions.  Additionally, this is an infinite space; therefore, the game need not have an equilibrium at all, according to the theory of the existence of a Nash equilibrium as discussed by \cite{nash1950equilibrium}. So the stability issues make it challenging to train a GAN to perform regression. The authors of Deep Regret Algorithm GAN (DRAGAN) use Sion's Theorem to counter the hypothesis that the stability issue comes from using a solid distance measure, and instead, they argue that since the GAN game is not necessarily convex-concave, the game may converge to undesirable local equilibria that result in mode collapse. The authors present a regularization term to navigate the map of local equilibria. This method also proves to be effective, but it does not universally solve the convergence issue either. (\cite{kodali2017convergence},\cite{fedus2017many})

These two competing theories from \cite{arjovsky2017wasserstein} and \cite{kodali2017convergence} dominate GAN stability, so it is essential to consider these theories when designing a GAN architecture.

The second problem inherent to GANs for regression is related to competing terms within the loss function. Regression GAN (Reg-GAN) was used to predict steering angles from street images, and it is similar to semi-supervised GAN (SGAN) (\cite{odena2016semi}), where the output labels are classes corresponding to the inputs. However, the regression problem requires an infinite number of classes because each input belongs to a different class. The Reg-GAN simplifies the regression problem by constructing the discriminator in a way where $N$, the number of classes, is enormous; the higher the $N$, the more accurate the model. If part of the loss function represents the GAN game and an additional term is included to represent the regression loss, then the regression performance introduces instabilities in the training of entire framework, the architecture simplifies to a DNN and the utility of two networks competing with each other is lost and the basic premise of the GAN framework is not leveraged. Therefore, Reg-GAN also adopted the concept of feature matching to stabilize the training. Feature matching is the concept of forcing the generator to match the output of each layer of the discriminator from the real input rather than just the final output (\cite{rezagholizadeh2018semi}). However, in Reg-GAN  the feature matching did not promote competition between the generator and discriminator and thus still introduced some instabilities so generalized semi-supervised regression GAN (SR-GAN) modified the loss function by introducing the concept of feature contrasting to compliment the feature matching. Feature contrasting promotes competition between the discriminator and generator by forcing the generator to match the output of a layer of the discriminator and forcing the discriminator to try and increase that distance. The SR-GAN is currently the highest functioning regression GAN (\cite{olmschenk2018generalizing}). However, the performance of SR-GAN on the mentioned problem sets is further improved through our proposed technique.  

Alternatively, the authors of (\cite{aggarwal2019benchmarking}) evaluated the efficacy of using conditional GAN (CGAN) for regression. They modified the architecture so that the generator receives two inputs: a noisy input and $x$. The inputs each feed into their fully connected layer before eventually being concatenated to be processed together to predict $y$. The authors compare the accuracy of CGAN against modern deep networks such as XGBoost, MDNs, and DNN. There wasn't an additional loss term so the regression was built into the architecture design--stability in training is directly related to improved regression performance.

Therefore, it is advantageous to design a network with a loss function that simultaneously guarantees the accomplishment of both goals. The network should perform regression to benefit stability and leverages a stable game to aid in the regression prediction.  Above mentioned challenges need to be addressed to adopt GAN architectures for the regression tasks.

\subsection{Solution}

To alleviate the problems mentioned earlier, we introduce a fuzzy logic system in multiple locations of the GAN architecture. We observe that adding a fuzzy logic system improves the stability of the training through our experimental results. Additionally, fuzzy logic has been used for regression applications (\cite{tamilarasan2015approach}, \cite{kovac2013application}, \cite{chukhrova2019fuzzy}) and has shown desirable results; so here we evaluate the results of applying a fuzzy logic system to GANs. The use of fuzzy logic in GANs is novel and could increase GAN game formulation's scope for stability and improve the performance of GANs for regression.

This paper embeds the fuzzy-logic based system in three ways, and we evaluate the efficacy of each embedding: (1) predicting the output, $y$, directly, (2) evaluating the performance of the generator, and (3) both the prediction of $y$ and the evaluation of the generator. In particular, the major contributions of the paper are enumerated as follows:

\begin{enumerate}
    \item The proposed architecture improves the regression capabilities of GANs. Besides, ensuring stability in the GAN game across the networks evaluated. The enhanced predictive capability of this new class of GANs, i.e., fuzzified ones, can be attributed to the three different methods of infusing fuzzy logic in a GAN architecture. The methods are as follows: \textit{Fuzzy regression injection}, \textit{Fuzzy classification injection}, and \textit{Fuzzy combined injection}. The details of the particular techniques are elaborated on in the following sections. 
    
    
    \item We evaluate the robustness of the proposed methods on five different yet illustrative datasets, and the empirical results stand witness to the superior regression capabilities of the fuzzy-infused GANs compared to traditional ones.





\end{enumerate}

The subsequent sections will discuss and evaluate the proposed novel injection techniques. Section 2 outlines the related work and includes an introduction to the proposed injection techniques. In Section 3, we develop a technical description of the injection techniques by outlining the mathematical theory behind fuzzy logic-based systems, their differentiability, and how it improves convergence. Following Section 4 introduces the GAN frameworks to be augmented and the problems used to evaluate CGAN. Section 5 discusses the methodology used to evaluate the performance of the proposed frameworks, and section 6 reports the results. Finally, section 7 revisits the topics discussed in sections 2-6 and concludes the paper.

\section{Methodology}

Instead of introducing a regularization penalty, we introduce an adjustment to the GAN game that incorporates the aggregation of fuzzy implications to improve the evaluation of the real and generated distributions. We will not compare the convergence or stability of the introduced formulation against modern techniques since this paper focuses on improving regression and evaluating GANs' ability to do so. Although stability will be necessary for performing and improving regression, evaluating the stability and performance of GANs on non-image data is still an open problem in literature (\cite{lucic2017gans}, \cite{grnarova2019domain},
\cite{fekri2020generating}). Due to the enormous scope, we leave the evaluation of the stability of the proposed GAN formulation open-ended to provide avenues for future research. Instead, we will provide insight into why we conclude that we observe the improvement of convergence and stability over the algorithms performing regression.

Secondly, fuzzy logic has accurately performed regression in the absence of accurate mathematical models. This theory has been experimentally demonstrated by \cite{chiu1997extracting},  \cite{kikuchi1994estimation}, and \cite{kovac2013application} in trip prediction, trip generation, and surface roughness modeling in face milling to be an effective means of regression through semantic variables.

This success can be attributed to the approximate reasoning of a noisy input with high uncertainty. GAN game played between $D$ and $G$ results in an inherent uncertainty in the outputs of $D$ or $G$ during training. The precarious nature of the outputs makes it challenging to apply GANs for regression; however, if we apply a fuzzy logic system to the output of the component in GAN that will perform the regression, the operations of the fuzzy logic system will be less dependent on the individual outputs of the network. The conclusion drawn by the fuzzy logic system will concern itself more with the accuracy of the aggregation of the entire output. Therefore, individual values do not have to be entirely accurate for the fuzzy logic system to develop an accurate approximation of the output, as long as the latent space is somewhat accurate as a whole. Thus reducing the need to learn downsampling to a single output, and the downsampling will occur within the fuzzy logic system.


\subsection{CGAN}
CGAN is not traditionally used for regression, but its regression capabilities were measured by \cite{aggarwal2019benchmarking}. We adopt their framework and apply the fuzzy logic to this algorithm. 

In the paper, CGAN is modified such that the generator receives inputs, $x$, and $z$, where $x$ is the real input, and $z$ is noise. The output of the generator is $\hat{y}$. So this architecture utilizes the generator for regression. CGAN does not include an additional loss term for the difference between $y$ and $\hat{y}$; instead, the network only uses the original formulation of the game (Equation \ref{eq: minimax}) as its loss function. The paper outlines different network structures for each dataset; however, we only use two of their network structures during the evaluation of the datasets because the data sets we chose only span two network structures. The hyperparameters for the network are listed in Appendix A.

\subsection{Benchmarking with DNN}
DNN is a network that feeds forward without any backward flow besides backpropagation. There are many classes of DNNs, and here we use a DNN composed of only fully connected layers. The structure and hyperparameters for training are shown in Appendix A.


\subsection{Fuzzy Logic}
Fuzzy logic is an abstraction of classical logic where the potential outputs are real numbers in [0, 1]; this varies drastically from the binary output of classical logic--the output is only either true or false. Fuzzy logic models the concept of vagueness by arguing that the truth value of many propositions can be noisy to measure. There are two types of fuzzy logic: propositional and predicate. Propositional fuzzy logic consists of fuzzy conjunctions, disjunctions, negations, and implications; predicate fuzzy logic generalizes propositional fuzzy logic with universal and existential quantification. We will focus on predicate fuzzy logic systems since it is an extension of propositional fuzzy logic. Recently a subset of predicate fuzzy logic was introduced: differentiable fuzzy logic (DFL). DFL consists of all functions of predicate fuzzy logic that are differentiable; this is desirable as differentiability is necessary for functions in a deep learning algorithm.

\subsection{Differentiable Fuzzy Logic}
\label{DFL}
Differentiable fuzzy logic (DFL) is a subclass of predicate fuzzy logic composed of differentiable operators for computing the fuzzy logic interpretation. \cite{van2020analyzing} evaluates the different differentiable fuzzy logic operators and demonstrates the shortcomings of certain operators when applied to deep learning. Most operators do not have intuitive derivatives that correspond to inference rules from classical logic. However, the paper mentions that those properties hold for product-based operators. We present the product-based differentiable fuzzy logic operators and the related terminologies in the subsequent subsections to develop the operator, $Fuzzy(\cdot)$, used throughout the proposed injection techniques in Section \ref{Injection}.

\subsubsection{Semantics}
Additionally, DFL defines new semantics, which extends traditional semantics to use vector embeddings.  A structure in DFL again consists of a domain of discourse and an embedded interpretation:
\begin{definition}

A DFL structure is a tuple $S = \langle O, \eta, \theta \rangle$, where $O$ is a finite but unbounded
set called domain of discourse and every $o \in O$ is a d-dimensional vector, $\eta : P \times \mathbb{R}^W \rightarrow (O^m \rightarrow [0, 1])$ is an
(embedded) interpretation, and $\theta \in \mathbb{R}^W$ are parameters. $\eta$ maps predicate symbols $p \in P$ with arity $m$ to a
function of $m$ objects to a truth value $[0, 1]$. That is, $\eta(p, \theta) : O^m \rightarrow [0, 1]$. 
\end{definition}

That is, objects in DFL semantics are $d$-dimensional vectors of reals. The vectorized semantics are then used in the t-norm and t-conorm operators to resolve the network's truthness.

\subsubsection{Triangular Norms
}

In propositional and predicate fuzzy logic, the functions used to compute the conjunction of two truth values are called t-norms.

\begin{definition}
A t-norm (triangular norm) is a function $T: [0, 1]^2 \rightarrow [0, 1]$ that is commutative and associative. For all $a \in [0, 1]$, whenever $0 \leq b1\leq b2 \leq 1$, then $T(a, b1) \leq T(a, b2)$. and for all $a \in [0, 1], T(1, a) = a$.

\end{definition}

Following the definition above, many equations satisfy those conditions for a differentiable t-norm. However, this paper does not present a review of the algorithms presented in \cite{van2020analyzing}. We leverage the work of the cited paper to validate our choice of operators, as we discussed above. The t-norm that we use is the product t-norm:

\begin{equation}
    T(a,b) = a \circ b.
\end{equation}

Where we use the Hadamard product to do element-wise multiplication since $a$ and $b$ are vectors in our formulation. Therefore, $T(a,b)$ is a vector with the same shape of $a$.

\subsubsection{Triangular Conorms}

Similarly, the functions used to compute the disjunction of two truth values are called t-conorms or s-norms.

\begin{definition}
A t-conorm  is a function $S: [0, 1]^2 \rightarrow [0, 1]$ that is commutative and associative. For all $a \in [0, 1]$, whenever $0 \leq b1\leq b2 \leq 1$, then $S(a, b1) \leq S(a, b2)$ and for all $a \in [0, 1]$, $S(0, a) = a$.

\end{definition}

The t-conorm corresponding to the product norm is shown in Equation \ref{t_conorm}.

\begin{equation}
    S(a,b) = a + b - a \circ b.
\label{t_conorm}
\end{equation}
Similar to the t-norm we use the Hadamard product so $S(a,b)$ is also a vector with the same shape of $a$.

\subsubsection{Fuzzy Implications}
\label{FuzzyImp}
The functions that are used to compute the implication of two truth values are called fuzzy implications.
\begin{definition}

 A fuzzy implication is a function $I : [0, 1]^2 \rightarrow [0, 1]$ so that for all $a, c \in [0, 1]$, $I(·, c)$ is
decreasing, $I(a, ·)$ is increasing and for which $I(0, 0) = 1, I(1, 1) = 1$ and $I(1, 0) = 0.$
\end{definition}

Similarly, many algorithms satisfy the above definition that is differentiable implications. However, we use a product-based implication, the Reichenbach implication:

\begin{equation}
    I_{RC}(T,S) = 1 - T + T \circ S.
\end{equation}

To increase gradients for values of the consequent that make the implication less accurate, we apply an additional sigmoidal function to the implication that yields a sigmoidal implication, which is also an implication \cite{van2020analyzing}. 

\begin{equation}
    I(T,S) = \frac{(1 + e^{9 / 2}) \cdot \sigma(9(I_{RC}(T,S) - 1/ 2)) - 1}{e^{9/2} - 1}.
\end{equation}

Where $\sigma(\cdot)$ is the sigmoid function. 

\begin{equation}
    \sigma(x) = \frac{1}{1 + e^{-x}}.
\end{equation}

\subsubsection{Aggregation}

Lastly, from predicate fuzzy logic, functions used to compute quantifiers like $\forall$, and $\exists$ are represented as aggregation operators.
 \begin{definition}
 
 An aggregation operator is a function $A : [0, 1]^n \rightarrow [0, 1]$ that is symmetric and increasing with respect to each dimension, and for which $A(0, ..., 0) = 0$ and $A(1, ..., 1) = 1$. A symmetric function is one
in which the output value is the same for every ordering of its arguments.
 \end{definition}

Lastly is the aggregation step, which combines all of the implications to make a final prediction. The aggregation function used is the product aggregator. This aggregator is expressed in Equation \ref{agg}.

\begin{equation}
    \label{agg}
    A(x_1,...,x_n) = \prod_{i=1}^n x_i.
\end{equation}

Where $x$ is the substitute for the implications, $I$.




\subsubsection{DFL Implementation Approaches}
\begin{figure}[H]
    \centering
    \includegraphics[width= 5in]{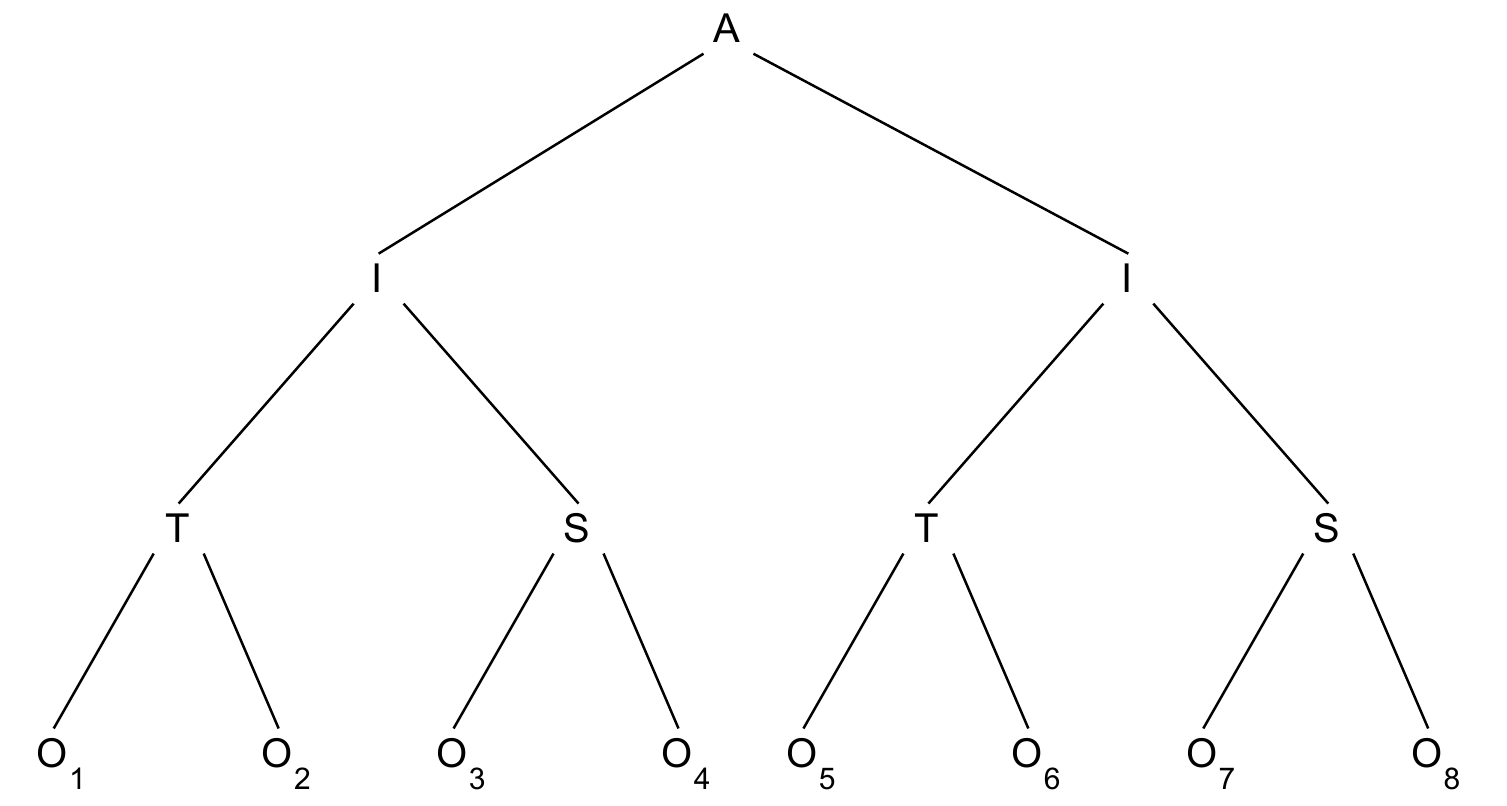}
    \caption{Example Fuzzy Logic Structure}
    \label{fig:ex_fuzzy}
\end{figure}

In Figure \ref{fig:ex_fuzzy}, $O_N$ represents the outputs of the network, in other words, we modify the output of either the discriminator or generator to output a vector rather than a scalar value, $T$ represents a t-norm, $S$ represents a t-conorm, $I$ represents an implication, and $A$ represents an aggregation. This figure is used to outline the flow of information through the fuzzy logic system. As we discuss the different injection techniques, we want to preserve how the fuzzy logic injection behaves.

\subsection{Fuzzy Logic Injection}
\label{Injection}
\subsubsection{Regression Injection}
\begin{figure}[H]
    \centering
    \includegraphics[width= 5in]{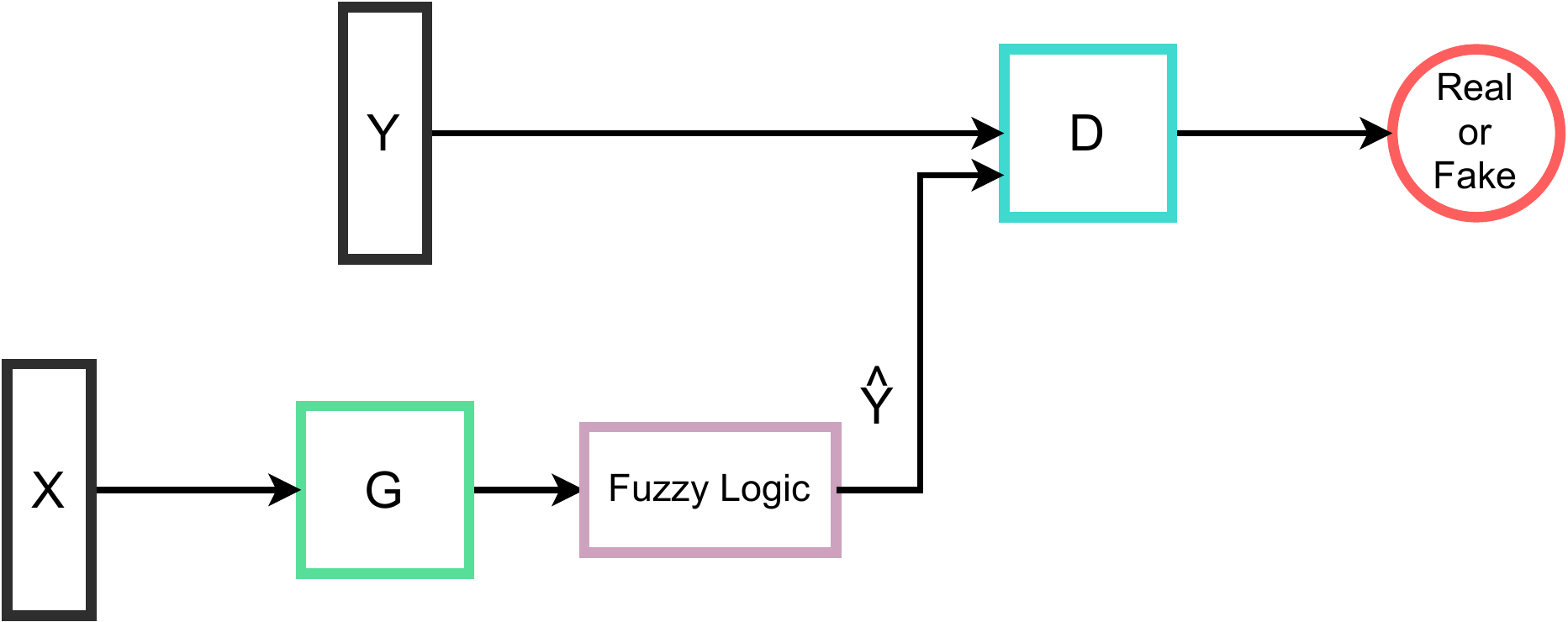}
    \caption{Regression Injection: Generator}
    \label{fig:Reg_inj_2}
\end{figure}

The regression injection technique (FRI) is employed to inject the fuzzy logic to the output of the GAN that controls the prediction of $Y$. The method could be introduced in either the generator or the discriminator, so the models are shown in Figure \ref{fig:Reg_inj_2}.


For this fuzzy logic system, the output of the GAN corresponding to the regression will be an $N$-dimensional vector where $N$ is the number of abstract features the network has to identify. Therefore, if the discriminator is performing the regression, instead of a single output, the output will be an $N$-dimensional vector, and in the other scenario where the generator would typically output a scalar value to perform the regression, the output would instead be the $N$-dimensional vector. These values represent a probability, $P$,  of those features existing. The $N$-dimensional output will be partitioned into four vectors: $a^{j \times 1}$, $b^{k \times 1}$, $c^{l \times 1}$, and $d^{m \times 1}$. These four vectors can vary in size as long as $j + k + l + m = N$.  The size of the four vectors is a hyper-parameter to be tuned. $a^{j \times 1}$ and $b^{k \times 1}$ are combined in a conjunction, $T$, and $c^{l \times 1}$ and $d^{m \times 1}$ are combined in a disjunction, $S$. $T$ is then used as the antecedent, and $S$ is used as a consequent in a fuzzy implication, $I$. If the length of the longest vector is: $M$, the shape of $I$ should be: [batch size, $M$]. Each column, $M$, represents an implication. The implications are aggregated over the columns to create an output vector equal to the size of the batch size. Through our experiments we found that $N=5$, $j=1$, $k=2$, $l=1$, and $m=1$ worked best; however, these sizes can be tuned as needed.

\subsubsection{Classification Injection}
\begin{figure}[H]
    \centering
    \includegraphics[width= 5in]{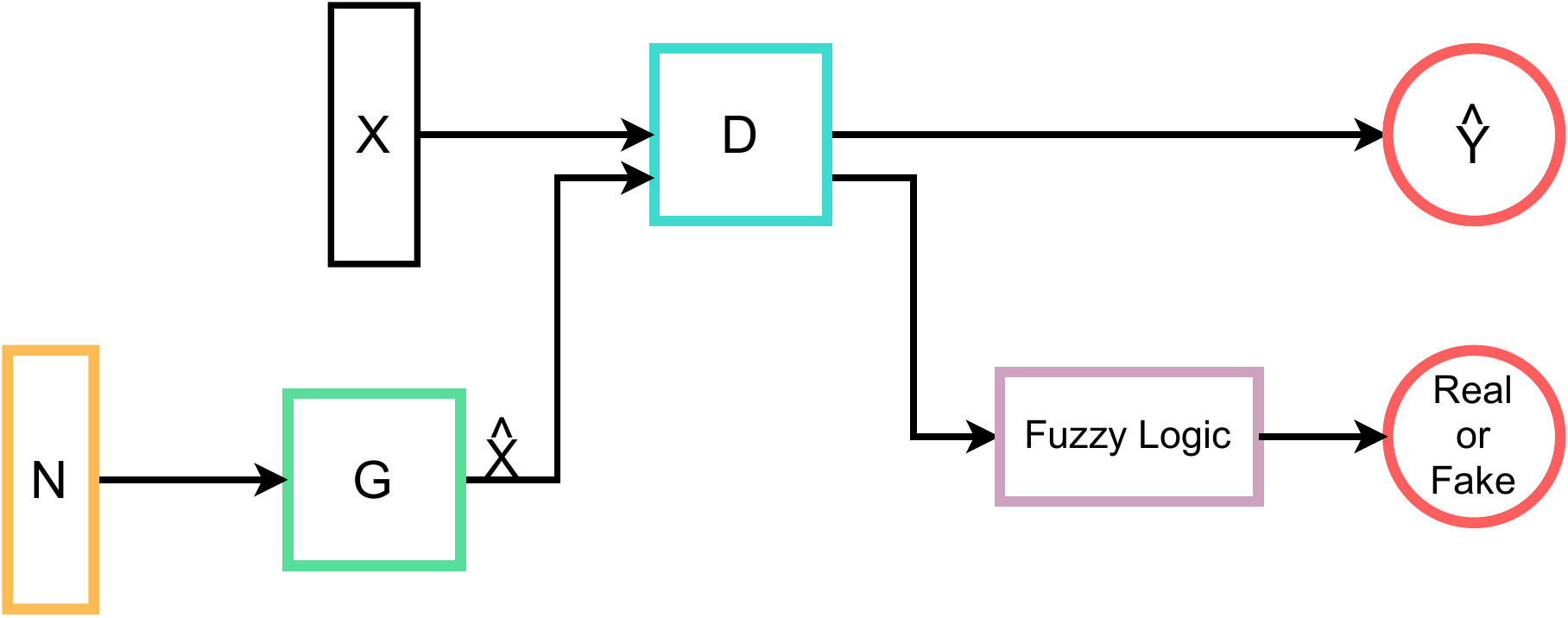}
    \caption{Classification Injection}
    \label{fig:Class_inj_1}
\end{figure}

The classification injection technique (FCI) is employed to inject the fuzzy logic to the output of GAN that controls the classification of real or fake data. The injection technique is shown in Figure \ref{fig:Class_inj_1}.

The procedure for extracting the truth value is similar to the regression injection technique, but the technique is applied at a different location. Similarly, through our experiments we found that $N=5$, $j=1$, $k=2$, $l=1$, and $m=1$ worked best for this injection as well.

Below we outline the modification to the GAN game through the classification injection using fuzzy logic. However, as mentioned above, we will not evaluate this method for improving stability on the GAN game; instead, we will only be looking at its effect on the regression performance of the GAN and conclude its stability improvement.

\begin{equation}
Fuzzy(\cdot) = A(I(T(a^*,b^*),S(c^*,d^*))).
\end{equation}

Where $A$ is the aggregation of fuzzy implications, $I$ is the vector of fuzzy implications, $T$ is the vector of t-norms, $S$ is the vector of t-conorms, and $a^*$, $b^*$, $c^*$, and $d^*$ represent the vectors $a$, $b$, $c$, and $d$, but modified so that the size of $a^*$ is equivalent to the size of $b^*$ and similarly for $c^*$ to $d^*$. The functions used are outlined in Section \ref{DFL}.

\begin{equation}
         V(G,D) = E_{z \sim p_{z}}[log(1 -Fuzzy(G(z)))] +  E_{x \sim p_{data}}[log(Fuzzy(D(x)))].
\end{equation}






\subsubsection{Double Injection}
\begin{figure}[H]
    \centering
    \includegraphics[width= 5in]{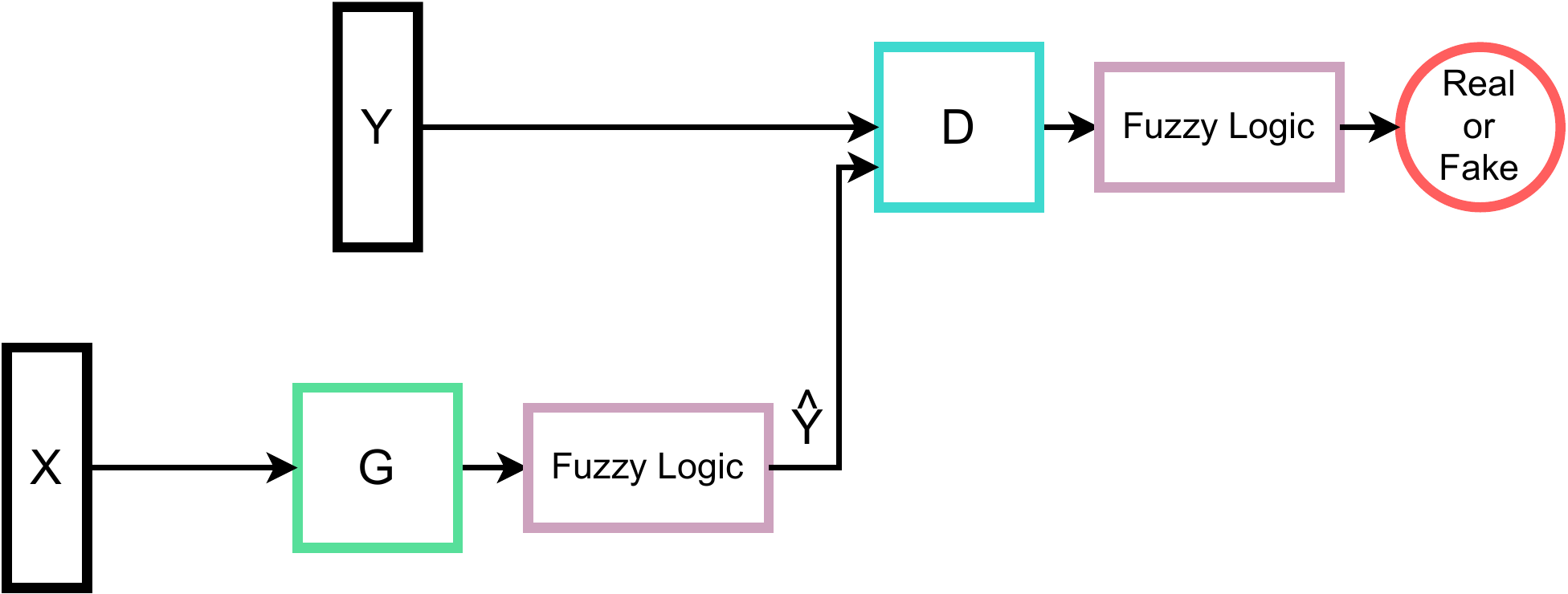}
    \caption{Double Injection: Generator}
    \label{fig:Com_inj_2}
\end{figure}

The double injection technique (FDI) employs both the regression injection and classification injection. The adopted methodology helps to conclude the combined deployment. The injection technique is shown in Figure \ref{fig:Com_inj_2}.


In the figure above, $Nz$ represents random noise, $X$ represents the ground truth input, $\hat{X}$ represents the predicted value of $X$, $Y$ represents the ground truth output, $\hat{Y}$ represents the predicted output, $D$ represents the discriminator and $G$ represents the generator.


\subsection{Evaluation}
The regression will be evaluated using both MSE (Equation \ref{MSE}) and MAE (Equation \ref{MAE}) metrics. The metrics use the difference between the real output, $Y$, and the predicted output, $\hat{Y}$.

\begin{equation}
    MSE = \frac{1}{n} \sum_{j=1}^{n}{(Y_j - \hat{Y}_j)^2}.
    \label{MSE}
\end{equation}

\begin{equation}
    MAE = \frac{1}{n} \sum_{j=1}^{n}{|Y_j - \hat{Y}_j|}.
    \label{MAE}
\end{equation}






\section{Data sets}

The data sets used are broken into two categories: non-image (\cite{Dua:2019}) and image (\cite{olmschenk2018generalizing}) data. We examine the efficacy of our method applied to various GAN architectures designed to operate on different data types. We adopt the data sets from their papers to provide a fair comparison.

\subsection{Data}
\label{Dataset_Non}
\textit{Abalone:} The age of abalone is dependent on the number of rings on its shell. Determining the number of rings requires cutting the shell through the cone, staining it, and counting the number of rings through a microscope. This process is time-consuming, which motivates the necessity for predictions based on data. This data set comprises 4177 samples with eight features: sex, length, diameter, height, total weight, shucked weight, viscera weight, and shell weight to predict the number of abalone rings. The justification for using this data set is that it develops a relationship from physical measurements to a physical characteristic of a living creature. We remove the sex feature to reduce our inputs to seven and make predictions from those seven other features.
 
\textit{Ailerons:} Flying an F16 requires the pilot to make rapid decisions based on the status of the aircraft at a particular time. The control inputs of the pilot must be precise to prevent disaster. These disasters can mean life or death. Due to the severity of these consequences, we must reduce the human error probability introduced to the system. Ideally, this requires the removal of the human element and automating the process of piloting an F16. Doing so requires data about the status to decide on the control action. This data set is composed of 13750 samples and 40 continuous inputs. Inputs 25-38 provide very little information about the status of the aircraft. Those values are almost all 0 for the entire data set. Due to this invariability, we considered removing the 14 samples, but the difference in performance was negligible, so we kept all 40 features. The control action's unnormalized output is scaled by 10,000, so the results are easier to comprehend. The justification for using this data set is that it shows how accurately the algorithms can assess a situation and control a complex system.

\textit{Bank:} A successful business understands how their customers behave given unsatisfactory conditions. In a bank, it is unsatisfactory for the customer to wait due to a full queue. So predicting whether these customers will leave or stay will help the banks modify their operating conditions. This data set is created by synthetically generating data from a simulation of how bank customers choose their banks. Tasks are based on predicting the fraction of bank customers who leave the bank because of full queues.  Each bank has several queues that open and close according to demand. The object is to predict the fraction of customers turned away from the bank because all the open tellers have full queues. This data set is composed of 8192 samples and 32 continuous inputs. Due to the data generation process, the predictions are only reliable for predicting the behavior based on the simulation results. The percentage is scaled by 10. The justification for this data set is that it provides a model for predicting bank-customer behaviors.

\textit{Census.} Purchasing a house is a lifetime goal for many people; whereas, selling houses is a profitable business opportunity. This relationship between buyer-to-seller requires a proper evaluation of the median price of a house, so both parties are satisfied with the result. The family of census house data sets is each composed of different input sizes and types. The sizes are either 8 or 16, and the types are either low or high difficulty. The difficulty level represents the relationship between inputs to outputs. A high difficulty has a more complex relationship than a low difficulty type. We use the 16 input, high difficulty data set for our training. The median price is scaled by $10^{-5}$.
 
\textit{Pumadyn:} Calculating the angular acceleration of each robot arm link in a Pumadyn robot is usually done through many matrix multiplications. The justification for this data is to measure how well the algorithm performs at mapping a known relationship. This data set was generated by a realistic simulation of the dynamics of a Puma 560 robot arm. Rather than calculating the angular acceleration of each arm, the goal of this data set is to predict the angular acceleration of one of the robot arm's links. The data set comprises 8192 samples, and 32 inputs include the angular positions, velocities, and torques of the robot arm. We scale the angular accelerations by 1000.

\section{Results: Fuzzy Logic Injection Techniques}

The data is normalized using a min-max normalization technique, so the values are between 0 and 1, so the MAE and MSE results are normalized.

\label{LayerResults}

\subsection{Regression Injection Results}

\begin{table}[H]
    \centering
    \begin{adjustbox}{width=\columnwidth,center}
    \begin{tabular}{|c|c|c|c|c|c|}
    \hline
         Network & Dataset &  \multicolumn{2}{c|}{NMAE} & \multicolumn{2}{c|}{NMSE} \\
         \cline{3-6}
         & & With Fuzzy & Without Fuzzy & With Fuzzy & Without Fuzzy\\
         \hline
         \multirow{5}{*}{CGAN} & Aileron & $0.04295\pm$ $6.26E-5$  & $\mathbf{0.04224\pm}$ $\mathbf{1.21E-4}$ & $0.003263\pm$  $4.96E-5$  & $\mathbf{0.003261\pm}$ $\mathbf{7.68E-6}$  \\ \cline{2-6}
         & Abalone & $0.05700\pm$ $2.36E-4$ & $\mathbf{0.05624\pm}$ $\mathbf{2.78E-4}$ & $0.006734\pm$ $2.68E-5$ & $\mathbf{0.006344\pm}$ $\mathbf{4.05E-5}$\\
         \cline{2-6}
         & Bank &  $\mathbf{0.07305\pm}$ $\mathbf{9.53E-6}$ & $0.08038\pm$ $7.04E-6$ & $\mathbf{0.01269\pm}$ $\mathbf{1.74E-6}$ & $0.01397\pm$ $2.00E-6$ \\\cline{2-6}
         & Census & $\mathbf{0.03536\pm}$ $\mathbf{8.56E-4}$ & $0.04178\pm$ $1.55E-4$ & $\mathbf{0.005554\pm}$ $\mathbf{2.13E-5}$ & $0.007278\pm$ $4.44E-5$ \\\cline{2-6}
         & Pumadyn & $\mathbf{0.05495\pm}$ $\mathbf{1.358E-4}$ & $0.1242\pm$ $1.81E-5$  & $\mathbf{0.004694\pm}$ $\mathbf{1.66E-5}$ & $0.02471\pm$ $5.78E-6$ \\\cline{2-6}
         \hline
       \end{tabular}
    \end{adjustbox}
    \caption{CGAN: Regression Injection Results}
    \label{tab:CGANREG}
\end{table}

Referencing Table \ref{tab:CGANREG}, the results of using FRI for CGAN vary in performance across data sets. In the ailerons and abalone data sets, FRI negatively affects the NMAE and NMSE. However, for the bank, census, and pumadyn data sets, FRI improves in both metrics, with the most significant improvement being pumadyn which saw a $55.76\%$ improvement in the NMAE and an $81.00\%$ improvement in the NMSE. The negative impact on the first two data sets is minute compared to the performance improvement on the other data sets; we believe the increased error results from the network's difficulty extracting high-level features from the input data. As mentioned in Section \ref{Dataset_Non}, the ailerons data set appears to be incomplete; the data from inputs 24-38 have very minimal differences and are near or equal to 0. Similarly, the abalone data set only has seven total inputs. We believe that these data sets do not provide enough information for CGAN to extract high-level features from the data, which mitigates the impact of the fuzzy logic system.

\subsection{Classification Injection Results}

\begin{table}[H]
    \centering
    \begin{adjustbox}{width=\columnwidth,center}
        \begin{tabular}{|c|c|c|c|c|c|}
    \hline
         Network & Dataset &  \multicolumn{2}{c|}{NMAE} & \multicolumn{2}{c|}{NMSE} \\
         \cline{3-6}
         & & With Fuzzy & Without Fuzzy & With Fuzzy & Without Fuzzy\\
         \hline
         \multirow{5}{*}{CGAN} & Aileron & $\mathbf{0.04220\pm}$ $\mathbf{9.29E-5}$   &  $0.04224\pm$ $1.21E-4$ & $\mathbf{0.003020\pm}$ $\mathbf{1.27E-5}$  & $0.003261\pm$ $7.68E-6$  \\ \cline{2-6}
         & Abalone &$\mathbf{0.05367\pm}$ $\mathbf{2.97E-4}$ & $0.05624\pm$ $2.78E-4$ & $\mathbf{0.005851\pm}$ $\mathbf{3.32E-5}$\ & $0.006344\pm$ $4.05E-5$\\
         \cline{2-6}
         & Bank &  $\mathbf{0.07277\pm}$ $\mathbf{7.67E-6}$ & $0.08038\pm$ $7.04E-6$ & $\mathbf{0.01129\pm}$ $\mathbf{1.46E-6}$ & $0.01397\pm$ $2.00E-6$ \\\cline{2-6}
         & Census &$\mathbf{0.03988\pm}$ $\mathbf{1.87E-4}$ & $0.04178\pm$ $1.55E-4$ & $\mathbf{0.006088\pm}$ $\mathbf{3.46E-5}$ & $0.007278\pm$ $4.44E-5$ \\\cline{2-6}
         & Pumadyn & $\mathbf{0.09227\pm}$ $\mathbf{1.38E-5}$ & $0.1242\pm$ $1.81E-5$ & $\mathbf{0.01359\pm}$ $\mathbf{4.66E-6}$ & $0.02471\pm$ $5.78E-6$ \\\cline{2-6}
         \hline
       \end{tabular}

    \end{adjustbox}
    \caption{CGAN: Classification Injection Results}
    \label{tab:CGANCLASS}
\end{table}

According to Table \ref{tab:CGANCLASS}, the results of using FCI for CGAN improved performance in every data set; the bank and pumadyn data sets exhibit significant changes. Using FCI for CGAN the bank data set has a $9.468\%$ reduction in NMAE and a $19.18\%$ change in NMSE. The changes for NMAE and NMSE for the pumadyn dataset are $25.71\%$ and $45.00\%$, respectively. This method improved the results universally because CGAN performs regression in the generator. The regression prediction and the convergence of the GAN game are working harmoniously. Therefore, improvements to the game's stability should also improve the generator's ability to create data in the output domain, $y$, which we observe through this experiment. The experimental outcomes help us believe there must be a stability upgrade from this injection technique, further motivating the notion of exploring the stability characteristics of this game.

\subsection{Double Injection Results}

\begin{table}[H]
    \centering
    \begin{adjustbox}{width=\columnwidth,center}
            \begin{tabular}{|c|c|c|c|c|c|}
    \hline
         Network & Dataset &  \multicolumn{2}{c|}{NMAE} & \multicolumn{2}{c|}{NMSE} \\
         \cline{3-6}
         & & With Fuzzy & Without Fuzzy & With Fuzzy & Without Fuzzy\\
         \hline
         \multirow{5}{*}{CGAN} & Aileron & $0.04427\pm$ $1.07E-4$   &  $\mathbf{0.04224\pm}$ $\mathbf{1.21E-4}$ & $0.003095\pm$ $7.33E-6$   & $\mathbf{0.003261\pm}$ $\mathbf{7.68E-6}$ \\ \cline{2-6}
         & Abalone &$0.05925\pm$ $5.26E-4$ &  $\mathbf{0.05624\pm}$ $\mathbf{2.78E-4}$ & $0.006655\pm$ $4.16E-5$ & $\mathbf{0.006344\pm}$ $\mathbf{4.05E-5}$ \\
         \cline{2-6}
         & Bank &  $\mathbf{0.06939\pm}$ $\mathbf{1.06E-5}$ & $0.08038\pm$ $7.04E-6$ & $\mathbf{0.01147\pm}$ $\mathbf{2.13E-6}$ & $0.01397\pm$ $2.00E-6$ \\\cline{2-6}
         & Census &$\mathbf{0.03967\pm}$ $\mathbf{1.35E-4}$ & $0.04178\pm$ $1.55E-4$ & $\mathbf{0.006437\pm}$ $\mathbf{3.02E-5}$ & $0.007278\pm$ $4.44E-5$ \\\cline{2-6}
         & Pumadyn & $\mathbf{0.08714\pm}$ $\mathbf{2.05E-4}$ & $0.1242\pm$ $1.81E-5$ & $\mathbf{0.01234\pm}$ $\mathbf{5.80E-5}$  & $0.02471\pm$ $5.78E-6$ \\\cline{2-6}
         \hline
       \end{tabular}
    \end{adjustbox}
    \caption{CGAN: Double Injection Results}
    \label{tab:CGAND}
\end{table}  

Using FDI for CGAN improved performance across the same three data sets when using FRI. Comparing the results in Tables \ref{tab:CGANREG}, \ref{tab:CGANCLASS}, and \ref{tab:CGAND} the only data set where using FDI proved to be the best method was the bank data set. This is counter-intuitive as both FRI and FCI improved performance across the bank, census, and pumadyn data sets. So it would be expected that a more considerable improvement would be observed using both techniques. However, the techniques are not complementary. One technique does not mitigate the weaknesses of the other—the reason why we do not universally see the most remarkable improvement from this technique.

\subsection{Compiled Results}
\begin{table}[H]
    \centering
    \begin{adjustbox}{width=\columnwidth,center}
    \begin{tabular}{|c|c|c|c|c|c|c|}
    \hline
         Dataset & Injection (GAN) &  \multicolumn{2}{c|}{NMAE} & \multicolumn{2}{c|}{NMSE}  \\
\cline{3-6}
         & & CGAN & DNN & CGAN & DNN\\
         \hline
         Ailerons & Classification & $\mathbf{0.04220\pm}$ $\mathbf{9.29E-5}$ & $0.09292\pm 2.55E-5$ & $\mathbf{0.003020\pm}$ $\mathbf{1.27E-5}$ & $0.09887\pm 7.12E-7$\\
         \hline
         Abalone & Classification & $0.05367\pm$ $2.97E-4$ & $\mathbf{0.05340 \pm 1.80E-6}$ & $\mathbf{0.005851\pm}$ $\mathbf{3.32E-5}$ & $0.006049 \pm 3.65E-7$\\
         \hline
         Bank & Double& $0.06939\pm$ $1.06E-5$ & $\mathbf{0.06384 \pm 8.66E-6}$ & $0.01147\pm$ $2.13E-6$ & $\mathbf{0.008151 \pm 3.11E-6}$\\
         \hline
         Census & Regression& $\mathbf{0.03536\pm}$ $\mathbf{8.56E-4}$ & $0.03618 \pm 2.54E-6$  & $0.005554\pm$ $2.13E-5$ & $\mathbf{0.005194 \pm 1.99E-6}$\\
         \hline
         Pumadyn & Regression & $\mathbf{0.05495\pm}$ $\mathbf{1.358E-4}$& $0.1337 \pm 4.45E-4$ & $\mathbf{0.004694\pm}$ $\mathbf{1.66E-5}$ & $0.02745 \pm 8.82E-5$ \\
         \hline
    \end{tabular}
    \end{adjustbox}
    \caption{Best Results vs. DNN}
    \label{tab:best}
\end{table}

\begin{table}[H]
    \centering
    \begin{adjustbox}{width=\columnwidth,center}
        \begin{tabular}{|c|c|c|c|c|c|c|}
    \hline
         Dataset & Injection (GAN) &  \multicolumn{2}{c|}{NMAE} & \multicolumn{2}{c|}{NMSE}  \\
\cline{3-6}
         & & CGAN & DNN & CGAN & DNN\\
         \hline
         Ailerons & Classification & $\mathbf{1.519\pm}$ $\mathbf{0.003342}$ & $3.345\pm 0.00115$ & $\mathbf{3.914\pm}$ $\mathbf{0.01641}$ & $128.1\pm 0.03312$\\
         \hline
         Abalone & Classification & $1.503\pm$ $0.00831$ & $\mathbf{1.501 \pm 0.00674}$& $\mathbf{ 4.588\pm}$ $\mathbf{0.0260}$ & $4.742 \pm 0.0225$\\
         \hline
         Bank & Double& $0.5892\pm$ $9.01E-5$ & $\mathbf{0.5421 \pm 7.87E-5}$ & $0.8271\pm$ $1.53E-4$ & $\mathbf{0.5877 \pm 4.77E-5}$\\
         \hline
         Census & Regression& $\mathbf{0.1768\pm}$ $\mathbf{4.28E-4}$ & $0.1809 \pm 3.44E-4$  & $0.1388\pm$ $5.33E-4$  & $\mathbf{0.1298 \pm 6.42E-4}$\\
         \hline
         Pumadyn & Regression &$\mathbf{8.241\pm}$ $\mathbf{0.0204}$& $20.06 \pm 0.0533$& $\mathbf{105.6\pm}$ $\mathbf{0.374}$ & $617.6 \pm 0.627$ \\
         \hline
    \end{tabular}

    \end{adjustbox}
    \caption{Unnormalized Errors}
    \label{tab:unnormalized}
\end{table}

Here (Tables \ref{tab:best}-\ref{tab:unnormalized}) we outline the best results for each data set and compare those results to a DNN. CGAN coupled with the different injection techniques improved the regression capabilities of the GAN architecture across the five non-image data sets tested in this paper. 
Using a regression injection, classification injection, or a double injection will depend on the data set. 

The fuzzy logic injected GANs are competitive with DNN across all five data sets. In the Abalone and Census datasets, the fuzzy logic injection improved the results of the specified GAN to either outperform the NMAE, NMSE, MAE, and MSE when the GAN without the injection was performing worse than DNN. Therefore, these techniques elevated GANs below a threshold of error that outperforms a DNN with comparable complexity, and this threshold was impassible by GANs before the introduced method.

\section{Conclusion}

This paper presents novel injection techniques that leverage a fuzzy logic system, explicitly using differentiable fuzzy operators, to improve a GAN architecture for regression. We show that the installment improves the tested GAN architecture's convergence while performing regression. The paper introduces three injection methods and compares those injection methods across various datasets.  The results show that these techniques will not universally improve regression. Instead, given the specific dataset, a desirable injection method optimizes regression potential. We show an injection method or multiple injection methods that improve the regression capabilities across the five datasets. The most substantial improvements were seen through the aileron and pumadyn data sets, which demonstrated a $54.58\%$ and $55.76\%$ improvement, respectively, in the NMAE. All other improvements were between 1\%-5\%.

These conclusions highlight three methods of leveraging fuzzy logic in the GAN architecture. However, we believe there are many avenues of improving the GAN game by introducing fuzzy logic. A re-framing of the GAN game as a fuzzy game could reduce existing local equilibria and ease the gradients that produce significant steps that overstep the optimal path to a desirable equilibrium. The GAN game framework is inherently unstable, so fundamentally, the framework needs to be modified to alleviate these issues. We conclude that fuzzy logic aids in improving GANs' ability to perform the regression, and we observe there must be a stability improvement across the datasets and networks tested; however, due to the lack of an evaluation of non-image data, we rely on the results of the performed regression. 

We propose pivoting the focus of researchers in developing a stability analysis of the proposed changes to the GAN game. We also believe there are many ways to incorporate fuzzy logic that are not outlined in this manuscript. Exploring those avenues could provide innovative solutions to improving the regression results. Additionally, we would like to explore avenues of selecting the features to be identified and passed to the fuzzy logic model--opposed to the network determining that on its own. Lastly, we will be studying the effects of modifying the aggregation of the implications in a way that tailors to a specified data set.


\bibliography{References}
\newpage

\appendix
\section*{Appendix A.}
CGAN Hyperparameters
\begin{table}[H]
    \centering
        \begin{adjustbox}{width=\columnwidth,center}
    \begin{tabular}{|c|c|c|}
          \hline     
          Hyperparameter & Generator & Discriminator  \\
        \hline
        Activation & elu & elu \\
        \hline
        Learning Rate & 0.001 & 0.001  \\
        \hline
        Learning Decay & 0.001 & 0 \\
        \hline
        Optimizer & Adam & Adam \\
        \hline
        Noise Input Size & 1 & Not Applicable \\
        \hline
         Number of Layers (before concat.) & 1 & 1\\
        \hline
        Size of Layers (before concat.) & 100 & 100 \\
         \hline
        Number of Layers (after concat.) & 5 & 4 \\
        \hline
        Size of Layers (after concat.) & 50 & 50 \\
         \hline
        Number of Epochs & 500 & 500 \\
        \hline
        Batch Size & 100 & 100 \\
        \hline
        Weight Initialization & He Normal & He Normal\\
        \hline
    \end{tabular}
        \end{adjustbox}
    \caption{Abalone}
    \label{tab:abaloneCGAN}
\end{table}

\begin{table}[H]
    \centering
        \begin{adjustbox}{width=\columnwidth,center}
    \begin{tabular}{|c|c|c|}
        \hline       
        Hyperparameter & Generator & Discriminator  \\
        \hline
        Activation & elu & elu \\
        \hline
        Learning Rate & 0.0001 & 0.0005  \\
        \hline
        Learning Decay & 0 & 0 \\
        \hline
        Optimizer & Adam & Adam \\
        \hline
        Noise Input Size & 1 & Not Applicable \\
        \hline
         Number of Layers (before concat.) & 1 & 1\\
        \hline
        Size of Layers (before concat.) & 100 & 100 \\
         \hline
        Number of Layers (after concat.) & 5 & 4 \\
        \hline
        Size of Layers (after concat.) & 50 & 50 \\
         \hline
        Number of Epochs & 500 & 500 \\
        \hline
        Batch Size & 100 & 100 \\
        \hline
        Weight Initialization & He Normal & He Normal\\
        \hline
    \end{tabular}
        \end{adjustbox}
    \caption{Ailerons}
    \label{tab:aileronsCGAN}
\end{table}
\begin{table}[H]
    \centering
        \begin{adjustbox}{width=\columnwidth,center}
    \begin{tabular}{|c|c|c|}
         \hline      
         Hyperparameter & Generator & Discriminator  \\
        \hline
        Activation & elu & elu \\
        \hline
        Learning Rate & 0.001 & 0.001  \\
        \hline
        Learning Decay & 0.001 & 0 \\
        \hline
        Optimizer & Adam & Adam \\
        \hline
        Noise Input Size & 1 & Not Applicable \\
        \hline
         Number of Layers (before concat.) & 1 & 1\\
        \hline
        Size of Layers (before concat.) & 100 & 100 \\
         \hline
        Number of Layers (after concat.) & 6 & 5 \\
        \hline
        Size of Layers (after concat.) & 100-75-75-75-75-50 & 50-50-50-50-25 \\
         \hline
        Number of Epochs & 500 & 500 \\
        \hline
        Batch Size & 100 & 100 \\
        \hline
        Weight Initialization & He Normal & He Normal\\
        \hline
    \end{tabular}
        \end{adjustbox}
    \caption{Bank}
    \label{tab:bankCGAN}
\end{table}

\begin{table}[H]
    \centering
        \begin{adjustbox}{width=\columnwidth,center}
    \begin{tabular}{|c|c|c|}
        \hline       
        Hyperparameter & Generator & Discriminator  \\
        \hline
        Activation & elu & elu \\
        \hline
        Learning Rate & 0.001 & 0.001  \\
        \hline
        Learning Decay & 0.0001 & 0 \\
        \hline
        Optimizer & Adam & Adam \\
        \hline
        Noise Input Size & 1 & Not Applicable \\
        \hline
         Number of Layers (before concat.) & 1 & 1\\
        \hline
        Size of Layers (before concat.) & 100 & 100 \\
         \hline
        Number of Layers (after concat.) & 5 & 4 \\
        \hline
        Size of Layers (after concat.) & 50 & 50 \\
         \hline
        Number of Epochs & 500 & 500 \\
        \hline
        Batch Size & 100 & 100 \\
        \hline
        Weight Initialization & He Normal & He Normal\\
        \hline
    \end{tabular}
        \end{adjustbox}
    \caption{Census}
    \label{tab:censusCGAN}
\end{table}
\begin{table}[H]
    \centering
        \begin{adjustbox}{width=\columnwidth,center}
    \begin{tabular}{|c|c|c|}
          \hline      
          Hyperparameter & Generator & Discriminator  \\
        \hline
        Activation & elu & elu \\
        \hline
        Learning Rate & 0.001 & 0.001  \\
        \hline
        Learning Decay & 0.001 & 0.001 \\
        \hline
        Optimizer & Adam & Adam \\
        \hline
        Noise Input Size & 1 & Not Applicable \\
        \hline
         Number of Layers (before concat.) & 1 & 1\\
        \hline
        Size of Layers (before concat.) & 100 & 100 \\
         \hline
        Number of Layers (after concat.) & 6 & 5 \\
        \hline
        Size of Layers (after concat.) & 100-75-75-75-75-50 & 50-50-50-50-25 \\
         \hline
        Number of Epochs & 500 & 500 \\
        \hline
        Batch Size & 100 & 100 \\
        \hline
        Weight Initialization & He Normal & He Normal\\
        \hline
    \end{tabular}
        \end{adjustbox}
    \caption{Pumadyn}
    \label{tab:pumadynCGAN}
\end{table}

The output layer is omitted from the discussion of the layers in the tables above. The output layer is modified from the original design as discussed in the paper.  The rest of the network design is held constant and those hyperparameters are displayed.

DNN Hyperparameters

\begin{table}[H]
    \centering
    \begin{tabular}{|c|c|}
    \hline
        Hyperparameter & DNN   \\
        \hline
        Activation & ReLU\\
        \hline
        Learning Rate & 0.001 \\
        \hline
        Learning Decay & 0.1 \\
        \hline
        Optimizer & Adam \\
         \hline
        Number of Layers & 7 \\
        \hline
        Size of Layers & 500-500-500-100-100-50-1 \\
         \hline
        Number of Epochs & 100 \\
        \hline
        Batch Size & 100 \\
        \hline
        Weight Initialization & Random \\
        \hline
    \end{tabular}
    \caption{Text Data DNN}
    \label{tab:textDNN}
\end{table}

Additionally, there is a dropout layer after each activation.

\end{document}